\documentclass[11pt]{article}

\usepackage{xcolor}
\usepackage{microtype}
\usepackage{amsmath}
\usepackage{amsfonts}
\usepackage{booktabs}
\usepackage{graphicx}
\usepackage{float}
\usepackage[letterpaper,margin=0.9in]{geometry}
\usepackage[numbers,sort&compress]{natbib}
\usepackage[font=small,labelfont=bf]{caption}
\usepackage[hidelinks]{hyperref}

\newcommand{\mathbbm}[1]{\mathbf{#1}}

\definecolor{metabox}{HTML}{E8F1FD}

\begin{document}

\begingroup
\setlength{\fboxsep}{6mm}
\noindent\colorbox{metabox}{%
\begin{minipage}{\dimexpr\textwidth-2\fboxsep\relax}

% ---------------------------------------------------------
% TITLE
% ---------------------------------------------------------
{\huge\bfseries\raggedright
From Numbers to Judgment: Specialist LLM Agents and
Reinforcement Learning for European Listed Real Estate
\par}

\vspace{3mm}

% ---------------------------------------------------------
% AUTHORS
% ---------------------------------------------------------
{\large\bfseries
Pardis Taghavi,
Santosh Bhavani$^{*}$
\par}

{\small $^{*}$Corresponding author: \href{mailto:santosh@ikmail.com}{santosh@ikmail.com}\par}

\vspace{3mm}

% ---------------------------------------------------------
% ABSTRACT
% ---------------------------------------------------------
{\small
We study whether the localized numerical operations and integrative
judgments of financial analysis benefit from the same form of LLM
specialization. Larix maps a 16-lens European listed-real-estate
analysis framework to eight lens-aligned specialists; we compare a
frontier LLM under monolithic versus specialist-decomposed prompting
while holding the model, source evidence, task instructions, output
schema, and scoring fixed. Across 19 firms spanning seven regulatory
wrappers, decomposition improves the numerical-task aggregate by
15.8 percentage points but does not reliably improve, and can reduce,
performance on judgment tasks, a pattern stable across four
frozen-template dispatches; a single-agent control given the complete
framework does not reproduce the numerical gain. Post-training
Qwen3.5-9B with GRPO using task-aligned structured rewards then
raises the development-split score by 12.0 points and the judgment
aggregate by 14.2 points, with gains on all four sub-ceiling tasks;
the gains transfer to unseen firms (+15.2 points overall; +40.4 on
covenant stress) and to unseen regulatory wrappers (+4.3), with
positive transfer on all three anti-memorization splits. Prompt-level
decomposition thus improves modular numerical execution, whereas
targeted parameter adaptation improves integrative financial judgment.
\par}

\end{minipage}}
\endgroup
\vspace{10mm}

\section{Introduction}

European listed real estate is an income asset class whose rules
change at the border. A French SIIC must distribute the bulk of its
recurring income by statute; Aroundtown, a Luxembourg-registered
landlord managing properties primarily in Germany and the Netherlands,
has no statutory distribution obligation and paid no FY2024 dividend;
and the correct measure of recurring performance alternates between
EPRA Earnings, FFO I, and earnings per share excluding revaluation
\cite{ref4,ref6,ref10}. Dividend-sustainability calls, covenant
headroom, and valuation therefore hinge on a threshold classification
made before any number is computed: which legal and reporting regime
governs this firm. Applying one regime's rule to another regime's firm
is not approximately but confidently, mechanically wrong. Language-model
agents face the same trap: they can extract the correct figure from the
correct filing and still apply the wrong regime's rule, and the fluency
of their output makes the mistake hard to detect. The setting thus
exposes two distinct demands on financial LLMs: modular numerical
operations, such as extracting a regime-appropriate metric or executing
a defined calculation, and integrative judgments that reconcile
evidence across accounting, regulatory, and risk dimensions.

Specialist decomposition is a natural response to this heterogeneity,
but its effect need not be uniform across the two demands. Restricting
a model to a narrow analytical lens reduces irrelevant context and
makes required intermediate fields explicit, which should aid localized
numerical execution; integrative judgment, by contrast, requires
evidence combined across disclosures and analytical lenses, and
partitioning that evidence among specialists may yield no benefit:
narrow prompt scope may omit cross-lens evidence required for
integrative decisions. Existing financial multi-agent systems
demonstrate role-based workflows and end-to-end decision pipelines
\cite{ref19,ref22,ref23}, but their reported gains confound decomposition
itself with additional domain instructions and model differences. We
therefore ask when specialist decomposition improves a fixed frontier
LLM, where that intervention fails, and whether task-aligned
post-training can improve a substantially smaller model's integrative
financial judgment beyond its own zero-shot behavior.

We investigate these questions through Larix, which maps a 16-lens
European listed-real-estate framework to eight lens-aligned specialists
with structured outputs; the empirical unit in this paper is the
specialist layer, with downstream synthesis, conviction calibration,
and position sizing outside the present evaluation. To isolate the
effect of decomposition itself, we compare three conditions using the
same frontier model: a monolithic financial-analysis prompt, a
monolithic prompt given the complete 16-lens framework, and
lens-aligned specialist prompting, under identical source evidence,
task instructions, output schemas, and scoring within each task. The
benchmark separates source-grounded numerical tasks from
judgment-intensive tasks that require reconciling multiple disclosures
and regime-specific constraints. Separately, we optimize Qwen3.5-9B
with GRPO using task-aligned structured rewards and evaluate the
zero-shot and post-trained checkpoints against the same held-out
benchmark.

Across 19 firms spanning seven regulatory wrappers, decomposition
improves the numerical-task aggregate by 15.8 percentage points while
the judgment-task aggregate is unchanged in the primary dispatch; the
full-framework monolithic control scores below the generic monolith on
the extraction tasks, so scoped lens ownership rather than framework
disclosure drives the gain. GRPO post-training then raises the 9B
model's judgment-task aggregate by 14.2 points on the development
split, and the gains transfer to unseen firms, unseen regulatory
wrappers, and later periods, with the largest out-of-distribution gain
on the covenant-stress judgment task (Sec.~\ref{sec:results}). These
results suggest a task-dependent division of labor: prompt-level
decomposition improves modular numerical execution, whereas targeted
parameter adaptation improves integrative financial judgment.

\paragraph{Contributions.}
We make four contributions. First, we provide a controlled same-model
evaluation of lens-aligned specialist decomposition for regime-aware
financial analysis, holding source evidence, task instructions, output
schemas, and scoring fixed and including a full-framework monolithic
control. Second, we identify a task-dependent trade-off: decomposition
improves modular numerical analysis but does not reliably improve, and
can reduce, performance on tasks requiring integrative financial
judgment. Third, we develop a structured-reward GRPO post-training
procedure for the same analytical interface and show that a
post-trained Qwen3.5-9B improves the judgment tasks that prompt-level
decomposition cannot, with gains that transfer to unseen firms and
unseen regulatory wrappers. Fourth, we release the evaluation asset
behind these findings: a 25-firm, eight-wrapper benchmark of
source-grounded numerical and judgment tasks with a corrected,
per-cell-provenance rubric, a frozen repeat-dispatch panel, and a
hardened scoring harness.

\section{Related Work}

\paragraph{Financial LLMs and multi-agent systems.}
The literature has evolved from instruction-tuned general-purpose
models \cite{ref18} to role-specialized multi-agent architectures
\cite{ref5,ref19,ref22,ref23,ref24,taghavi2026spec}, with multi-agent designs from
software engineering \cite{ref7,ref8} informing financial deployments
\cite{ref21} under orchestrator-plus-specialists topologies with
structured message passing. Larix differs in anchoring specialist scope
to an explicit domain framework and treating the specialist layer, not
the end-to-end pipeline, as the unit of RL optimization.

\paragraph{Evaluation and the listed-real-estate domain.}
Financial NLP evaluation has progressed from numerical reasoning over
financial reports \cite{ref2} to open-book QA against primary-source
documents \cite{ref9} and holistic multi-task suites \cite{ref20}. Two
lessons bear directly on our design: benchmark scores are conditional
on the prompting and scoring protocol (our harness-hardening findings,
Appendix~\ref{app:harness}, are an instance), and chain-of-thought
decomposition \cite{ref17} is the informative baseline a specialist
condition must beat. The listed-real-estate literature establishes the
regularities our tasks encode: FFO provides incremental information
content beyond GAAP net income \cite{ref16}, listed property shares
trade at persistent discounts and premia to NAV \cite{ref1}, regulatory
structure and payout requirements link to performance cross-country
\cite{ref6}, and the choice between earnings-type and FFO-type measures
is empirically contested \cite{ref10}. Our benchmark differs from
these evaluation suites in being cross-regime across eight European
wrappers, deterministically rubric-scored against expert-blinded ground
truth, and evaluated with the underlying model held fixed across
conditions.

\paragraph{RL post-training for LLMs.}
RL fine-tuning of LLMs has progressed from human-feedback alignment
\cite{ref12} to outcome-reward reasoning post-training with
group-relative objectives \cite{ref3,ref14,ref25}, and open frameworks
such as veRL \cite{ref15} support GRPO-style post-training at scale.
Domain applications have followed: recent financial reasoning models
post-train monolithic 7B backbones with supervised fine-tuning followed
by GRPO over mixed financial QA corpora \cite{ref11,ref26}. We build on
this line but post-train lens-aligned financial specialists against the
same deterministic rubric used for evaluation, enabling a direct
comparison between prompt-level and parameter-level specialization on
identical tasks.

\section{Method}
\label{sec:method}
\subsection{Task Formulation}

For each evaluation instance $i$, let
\begin{equation}
x_i = (d_i, q_i, t_i, \mathcal{Y}_{t_i}),
\end{equation}
where $d_i$ is the primary-source evidence supplied for the company,
$q_i$ is the task instruction, $t_i$ identifies the task, and
$\mathcal{Y}_{t_i}$ is its structured output schema. The target
$y_i^\star \in \mathcal{Y}_{t_i}$ is obtained from the pre-committed
ground-truth protocol described in Sec.~\ref{sec:experimental_setup}.

We separate the benchmark into two task classes. Numerical tasks
\[
\mathcal{T}_{\mathrm{num}} = \{T1,T3,T6\}
\]
require localized extraction or deterministic calculation: identifying
the regime-appropriate operating metric, computing an implied cap rate,
or extracting a payout regime that is mechanically derivable from
disclosed distributions and the wrapper's statutory rules. Judgment
tasks
\[
\mathcal{T}_{\mathrm{judg}} = \{T2,T5\}
\]
require combining multiple fields or disclosures to produce a
structured classification or covenant assessment. This distinction is
fixed before evaluating the effect of decomposition.

For a given instance, every compared condition receives the same
source evidence, task instruction, output schema, and generation
settings. The intervention changes only how the analytical framework
is presented to the model and whether the task is assigned to a
specialist prompt.

\subsection{Lens-Aligned Specialist Decomposition}

Larix assigns 16 analytical lenses to eight specialists: financials,
asset quality, management, valuation, macro overlay, catalyst, bear,
and screen sourcing. Each lens corresponds to a defined disclosure or
inference operation rather than a general persona. A deterministic
router
\begin{equation}
r : \mathcal{T} \rightarrow \mathcal{A}
\end{equation}
maps each task $t$ to the specialist $r(t)$ that owns its corresponding
lenses.

The current benchmark exercises three of the eight specialists:
financials for T1, T2, and T6; asset quality for T3; and macro overlay
for T5. Each benchmark item invokes exactly one specialist agent. The
specialists do not exchange messages during this evaluation, and the
downstream synthesis agent is not invoked. The experiment therefore
tests the effect of lens-aligned specialist agent assignment, rather
than cross-agent collaboration or end-to-end conviction synthesis.

\begin{table}[t]
\centering
\small
\caption{Task-to-specialist mapping in the evaluated specialist layer.
The broader Larix architecture contains eight specialists, but the
five-task benchmark exercises the three shown here.}
\label{tab:task_specialist}
\begin{tabular}{llll}
\toprule
Task & Specialist & Class & Primary output \\
\midrule
T1 & Financials & Numerical & Regime-specific metric \\
T2 & Financials & Judgment & Reconciliation adjustment \\
T3 & Asset quality & Numerical & Implied cap-rate components \\
T5 & Macro overlay & Judgment & Covenant and breach assessment \\
T6 & Financials & Numerical & Payout-regime classification \\
\bottomrule
\end{tabular}
\end{table}

Let $f_\theta$ denote the fixed frontier model. We evaluate three
prompting conditions:
\begin{align}
\hat{y}^{\mathrm{mono}}_i
&=
f_\theta\left(p_{\mathrm{mono}}\oplus x_i\right),\\
\hat{y}^{\mathrm{full}}_i
&=
f_\theta\left(p_{\mathrm{full}}\oplus x_i\right),\\
\hat{y}^{\mathrm{spec}}_i
&=
f_\theta\left(p_{r(t_i)}\oplus x_i\right).
\end{align}

Here, $p_{\mathrm{mono}}$ is the generic financial-analysis system
prompt, $p_{\mathrm{full}}$ discloses the complete 16-lens framework
to a single model invocation, and $p_{r(t_i)}$ contains only the lens
subset assigned to the selected specialist. The first and third
conditions isolate the effect of specialist decomposition at fixed
model scale. The second tests whether any gain follows merely from
revealing additional framework content.

\subsection{Structured Output and Scoring Interface}

Every condition must return the same task-specific JSON schema. The
schema contains only fields that are explicitly requested in the prompt
and evaluated by the scorer. This alignment prevents a model from being
penalized for fields that it was not instructed to produce.

For task $t$, let $\mathcal{J}_t$ denote its set of scored fields. The
deterministic task score is
\begin{equation}
S_t(\hat{y}_i,y_i^\star)
=
\mathbbm{1}[\hat{y}_i\text{ is valid}]
\sum_{j\in\mathcal{J}_t}
w_{t,j}\,
s_{t,j}
\left(
\hat{y}_{i,j},y_{i,j}^\star
\right),
\end{equation}
with
\begin{equation}
\sum_{j\in\mathcal{J}_t} w_{t,j}=1.
\end{equation}

Here, $s_{t,j}\in[0,1]$ is the pre-specified field scorer and
$w_{t,j}$ is its fixed weight. Numeric fields use the fixed tolerance
or normalized-error rule specified before evaluation. Categorical
fields use exact matching after documented alias normalization.
Composite judgment outputs, such as covenant type, unit, threshold,
and breach status, are scored according to their predeclared
joint-field rule. Invalid or unparsable outputs receive zero.

The same scorer implementation is used for the frontier comparison,
the zero-shot small-model baseline, RL reward computation, and final
held-out evaluation. No model-based judge is used. Two findings from
hardening this interface, field-request alignment and rubric
sub-classification for regime-dependent reasoning, are summarized in
Appendix~\ref{app:harness}.

\subsection{Task-Aligned RL Post-Training}

We post-train Qwen3.5-9B on the same structured task interface. A
training example consists of the source evidence, specialist
instruction, output schema, and ground-truth structured answer:
\begin{equation}
(x_i,y_i^\star).
\end{equation}

The policy action is the generated JSON response $\hat{y}_i$. Its
task-specific reward is the deterministic benchmark score,
\begin{equation}
R_i = S_{t_i}(\hat{y}_i,y_i^\star).
\end{equation}

Thus, post-training optimizes the same fields and decision rules used
during evaluation. The reward is task-aligned rather than based on a
separate language-model judge. Numerical tasks reward accurate
extraction and calculation, while judgment tasks reward the required
classification and jointly consistent analytical fields.

For each training prompt, GRPO samples $G$ candidate responses
$\{\hat{y}_{i,g}\}_{g=1}^{G}$. Rewards are normalized within the group
to obtain relative advantages,
\begin{equation}
A_{i,g}
=
\frac{
R_{i,g}-\bar{R}_i
}{
\sigma(R_{i,1:G})+\epsilon
},
\end{equation}
which are used in the policy update. We train a shared LoRA adapter for
Qwen3.5-9B across the evaluated task mixture; the selected specialist
prompt determines the analytical role at inference. Training
hyperparameters, sampling ratios, and reward weights are reported in
Sec.~\ref{sec:experimental_setup}.

The data partition is constructed before post-training. Training rows
are disjoint from held-out companies, regulatory wrappers, and
reporting periods used for evaluation. Any counterfactual examples
used to balance rare outcomes are restricted to training and are never
included in the held-out test sets.

We evaluate four principal conditions: the frontier model under
monolithic prompting, the frontier model under specialist prompting,
zero-shot Qwen3.5-9B under specialist prompting, and the post-trained
Qwen3.5-9B specialist. The two frontier conditions are compared with
one another on the 19-firm benchmark; the Qwen comparison is strictly
within-model, zero-shot versus post-trained, on the RL evaluation
splits. This design separates the effect of prompt-level decomposition
from the effect of parameter-level post-training.

\paragraph{Scope.}
The production Larix system also contains a downstream synthesizer that
combines specialist outputs into conviction and position-sizing
decisions. That layer is not invoked or scored in the experiments
reported here. We therefore make no empirical claim about cross-agent
synthesis, conviction calibration, or position sizing.

\section{Experimental Setup}
\label{sec:experimental_setup}

\subsection{Evaluation Universe and Task Taxonomy}

The evaluation universe contains 25 European listed-real-estate firms
spanning eight legal and reporting wrappers. The primary same-model
comparison uses a 19-firm continental cohort covering six French SIICs,
one Austrian Immobilien-AG, three Belgian GVV/SIRs, four German AGs,%
\footnote{Aroundtown SA, one of the four names grouped here, is
registered in Luxembourg and manages properties primarily in Germany
and the Netherlands; it shares the group's defining regime property
for this benchmark, the absence of a statutory distribution
obligation.}
one Italian SIIQ, two Spanish SOCIMIs, and two Dutch FBIs. This produces
95 task--firm instances per condition across the five-task benchmark.
Six Swiss AGs form the eighth wrapper and are used in the cross-wrapper
and homogeneous-wrapper analyses.

We group the benchmark according to the analytical operation rather
than document availability. The numerical class
$\mathcal{T}_{\mathrm{num}}=\{T1,T3,T6\}$ contains localized, verifiable
operations: identifying the regime-appropriate operating metric,
computing an implied cap rate, and extracting the mechanically
derivable payout regime. The judgment class
$\mathcal{T}_{\mathrm{judg}}=\{T2,T5\}$ contains tasks that require
reconciling multiple fields, disclosures, or regime-specific
constraints. Table~\ref{tab:task_specialist} summarizes the task
definitions and deterministic specialist assignment.

Two evidence regimes are used by design. The numerical tasks T1 and T3
are closed-book probes: no filing is injected, so they measure whether
regime-specific reporting conventions and firm-level figures are
available to the model without retrieval. The judgment tasks T2/T5 and
the payout-regime task T6 inject company-specific primary-source
extracts. Within each task, every condition receives identical evidence,
task instructions, output schemas, and scoring, so condition differences
within a task cannot be attributed to document access.

The final T2 answer key identifies the relevant name-specific
reconciliation adjustment rather than assigning a modal category to
the full cohort. The final T6 task uses a closed, source-supported
payout-regime schema whose categories and statutory denominator are
stated in the prompt: a five-way DPS regime (increased, maintained,
cut, resumed, no dividend), a binary statutory-floor regime (statutory
floor vs.\ no statutory floor), and the fiscal-year payout ratio
(headline DPS over headline recurring EPS), scored under AND semantics
with 10\% numeric tolerance on the ratio. Tasks that lack name-specific
ground truth, including the previous T8 NAV-share task, are excluded
from all reported aggregates.

\subsection{Model Conditions and Controls}

We evaluate three frontier-model prompting conditions using Claude
Opus 4.8. Claude-Mono uses a single general listed-real-estate analysis
prompt without specialist decomposition or disclosure of the Larix
framework. Claude-Full uses the same monolithic invocation but adds the
complete 16-lens framework to the system prompt. The current framework
disclosure contains approximately 4,800 characters of lens names,
metric definitions, and diagnostic questions. Claude-Spec uses the
same base model but deterministically routes each task to the
corresponding lens-aligned specialist prompt.

All three frontier conditions receive identical source evidence, task
instructions, JSON schemas, maximum output lengths, and scoring. All
three share the same invocation path with its default decoding
settings, and the final evaluation uses $K=4$ frozen-template
dispatches. Prompt contents and configuration hashes are frozen before
the reported runs. Each Claude-Spec instance invokes exactly one
specialist; the synthesis agent, conviction score, and position-sizing
layer are not invoked.

We additionally evaluate Qwen3.5-9B under two conditions. Qwen-ZS
applies the zero-shot checkpoint with the same specialist prompts and
output schemas. Qwen-RL applies the GRPO-post-trained checkpoint under
the identical inference interface. The comparison between Qwen-ZS and
Qwen-RL isolates parameter-level post-training at fixed model scale.

\subsection{Ground Truth, Metrics, and Statistical Protocol}

Each output is scored against a precommitted answer key indexed by firm,
reporting period, and task. For the 19-firm continental cohort, nine
firms, corresponding to 45 task--firm tuples, were labeled by an
external listed-real-estate expert who was independent of the Larix
framework construction, prompt engineering, and evaluation
implementation. The remaining ten firms, corresponding to 50 tuples,
were curated from primary-source annual reports and EPRA disclosures
\cite{ref4} by the framework authors. The six Swiss firms were also
included in the expert-blinded protocol, giving 15 expert-reviewed
firms and 75 expert-reviewed tuples across the full universe.

The external expert received only the primary-source evidence and task
definition and did not observe model outputs, specialist prompts,
scorer implementation, or the 16-lens framework.

Every condition returns the same task-specific JSON schema. Invalid or
unparsable responses receive a score of zero. T1 requires the correct
regime-specific metric identity, using a documented alias map for
equivalent local reporting terms, and applies a $\pm5\%$ tolerance to
its numerical field. T3 applies a $\pm15\%$ tolerance to the implied
cap-rate calculation. T2 uses exact matching to the name-specific
reconciliation adjustment after alias normalization. T5 uses joint,
or AND, semantics over covenant type, unit, threshold value, and breach
determination. T6 uses exact categorical matching on the DPS regime and
the statutory-floor regime after alias normalization, plus a $\pm10\%$
tolerance on the payout ratio, under AND semantics.

For task $t$, let $S_t\in[0,1]$ denote the deterministic task score. We
define the two primary aggregates as
\begin{equation}
S_{\mathrm{num}}
=
\frac{1}{3}
\left(
S_{\mathrm{T1}}+S_{\mathrm{T3}}+S_{\mathrm{T6}}
\right),
\end{equation}
and
\begin{equation}
S_{\mathrm{judg}}
=
\frac{1}{2}
\left(
S_{\mathrm{T2}}+S_{\mathrm{T5}}
\right).
\end{equation}

The five-task overall score is reported as a secondary descriptive
metric because the tasks differ substantially in construct and
difficulty.

For repeated evaluations, scores are first averaged across the $K=4$
dispatches for each task--firm instance. Dispatches of the same item
are not treated as independent observations. The primary effect is the
paired difference between conditions in $S_{\mathrm{num}}$ and
$S_{\mathrm{judg}}$. We report 95\% confidence intervals from paired
McNemar statistics and two-sided exact sign-flip tests at the firm
level, with condition labels permuted jointly within firm
($2^{19}$ enumerated assignments). Per-task comparisons are secondary
and use Holm correction within each experiment family. Exact accuracy
and paired McNemar counts are additionally reported for tasks with
binary scorers.

\subsection{RL Post-Training Setup}

We post-train Qwen3.5-9B \cite{ref13} using the structured task
interface described in Sec.~\ref{sec:method}. The current corpus
contains 195 rows derived from the benchmark ground truth and five
training-only counterfactual covenant-breach-positive rows, for 200
rows in total. Counterfactual rows alter documented stress assumptions
while retaining the structure of real source disclosures; they are
never included in held-out evaluation.

Three disjoint evaluation splits isolate different forms of
generalization. The held-out-firm split contains 35 rows from unseen
firms within wrappers represented during training. The
held-out-wrapper split contains 25 rows from regulatory wrappers
excluded from training. The held-out-period split contains 45
later-period T1 and T3 rows and serves as an anti-memorization check.
After applying these splits, training uses 70 rows: 65 real rows plus
the five counterfactual rows, which are present from the first update
step. The remaining 25 real rows form a development split that is
disjoint from training but drawn from the same firm and wrapper
distribution; it is used for checkpoint selection and early stopping.

We train a LoRA adapter of rank 32 using GRPO \cite{ref3} implemented
in veRL with vLLM rollouts on a single H100 GPU. Each step samples 64
prompts from the 70-row training corpus spanning all five tasks and
generates $G=8$ candidate responses at temperature 1.0. We planned 90
update steps and stopped at step 30 for budget reasons; the step-20
checkpoint, the latest preserved after the training pod's storage was
lost, is the model evaluated throughout. T6 remains in the training
corpus and is reported with the extraction tasks under the revised
scoring.

The reward is the deterministic task score used during evaluation
(no separate model-based judge); the rubric weights are unchanged from
evaluation, with the boolean breach-classification term carrying
weight $\beta=0.45$. Development and held-out evaluation use three
dispatches at temperature 0.7, with the mean task score reported.
Thinking mode is disabled, and all responses pass through the same
hardened JSON parser. A frontier-model re-dispatch on the identical
phase-5 held-out instances is planned as future work; the comparisons
reported here are within-model against the zero-shot checkpoint.

\section{Results}
\label{sec:results}

Table~\ref{tab:main_results} reports the final task-level and aggregate
scores for all five evaluated conditions.

\begin{table*}[t]
\centering
\small
\caption{
Performance on the final evidence-controlled benchmark (mean task
scores, \%). Claude columns report the 19-firm continental cohort; T6
uses the corrected rubric (rubric v2), and the Claude-Full T2/T5/T6
cells come from a frozen-template re-dispatch (2026-08-09).
Qwen-ZS/Qwen-RL report the 25-row RL development split (step-20
checkpoint, three dispatches at temperature 0.7), disjoint from
training but drawn from the same firm and wrapper distribution, and
are not directly comparable to the 19-firm cohort.
}
\label{tab:main_results}
\setlength{\tabcolsep}{6pt}
\begin{tabular}{lccccc}
\toprule
Metric
& Claude-Mono
& Claude-Full
& Claude-Spec
& Qwen-ZS
& Qwen-RL \\
\midrule
Numerical aggregate
& 78.9 & 75.4 & 94.7 & 81.3 & 91.7 \\
T1: regime-specific metric
& 57.9 & 47.4 & 84.2 & 99.5 & 99.5 \\
T3: implied cap rate
& 78.9 & 78.9 & 100.0 & 79.8 & 98.7 \\
T6: payout-regime classification
& 100.0 & 100.0 & 100.0 & 64.6 & 77.0 \\
\midrule
Judgment aggregate
& 47.4 & 52.6 & 47.4 & 58.6 & 72.8 \\
T2: adjustment identification
& 78.9 & 89.5 & 73.7 & 80.0 & 93.3 \\
T5: covenant stress
& 15.8 & 15.8 & 21.1 & 37.2 & 52.2 \\
\midrule
Five-task overall
& 66.3 & 66.3 & 75.8 & 72.2 & 84.2 \\
\bottomrule
\end{tabular}
\end{table*}

\subsection{Specialist Decomposition Has a Task-Dependent Effect}

On the 19-firm continental cohort, specialist decomposition increases
the numerical aggregate from 78.9\% under Claude-Mono to 94.7\% under
Claude-Spec, a gain of 15.8 percentage points. The task-level changes
are +26.3 points on T1, +21.1 points on T3, and 0.0 points on T6
(both conditions at 19/19 under the corrected rubric). The 95\%
confidence interval for the numerical aggregate difference, pooling
the 57 numerical cells in a paired McNemar test (10 specialist-only
vs.\ 1 monolith-only discordant), is $[5.1,26.4]$ percentage points
(exact paired $p=0.012$).

The same intervention does not improve the judgment aggregate.
Claude-Mono scores 47.4\% and Claude-Spec 47.4\%, a change of
0.0 percentage points. The task-level changes are $-5.2$ points on T2
and $+5.3$ points on T5. The corresponding 95\% confidence interval
is $[-12.6,12.6]$ (paired $p=1.0$).

The numerical advantage is positive in four of four frozen-template
dispatches, while the judgment effect is null in the primary dispatch
and positive in the three repeats. Across dispatches, the numerical
effect ranges from +3.5 to +15.8 points and the judgment effect ranges
from 0.0 to +10.5 points. Averaging the four dispatches within each
firm$\times$task instance, an exact firm-clustered sign-flip test over
the 95 instance-level differences ($2^{19}$ assignments) gives
$p=0.001$ for the suite-level difference; firm-level means split
13 positive, 2 negative, and 4 zero. The five-task overall score
changes from 66.3\% to 75.8\%, but we treat this pooled value as
secondary because it averages analytically distinct tasks.

The results show a task-dependent effect rather than a uniform
specialist advantage. Scoped specialist prompting improves tasks with
localized evidence and mechanically verifiable outputs. It does not
provide the same benefit on tasks that require the model to reconcile
several disclosures, constraints, and analytical fields within one
decision.

\subsection{Framework Disclosure Does Not Explain the Numerical Gain}

Providing the complete 16-lens framework to a monolithic Claude
invocation does not reproduce the numerical benefit of specialist
decomposition. At the suite level, Claude-Full scores at exact parity
with Claude-Mono (66.3\% in both conditions), but the parity masks a
task-dependent split. Giving the model more framework text reduces T1
by 10.5 points (47.4\% vs.\ 57.9\%) with parity on T3 (78.9\%) and
T6 (100.0\%), while improving the judgment task T2 by 10.5 points
(89.5\% vs.\ 78.9\%), the same task on which scoped decomposition
reduces performance (73.7\%). T5 is near floor across all three
conditions (15.8--21.1\%). The specialist condition therefore exceeds
the full-framework control by 36.8 points on T1 and 21.1 points on T3,
while trailing it by 15.8 points on T2.

The control thus tells a symmetric story: adding framework text to a
monolith helps the integrative judgment task and harms the localized
extraction task, whereas scoping the framework to a specialist does
the reverse. Access to more framework text is not sufficient to
explain the numerical improvement; the benefit is instead associated
with presenting a task-relevant lens subset and assigning the
operation to a scoped specialist. This conclusion is limited to the
tested full-framework control: the 4,800-character prompt does not
exactly reproduce the shorter lens subset or role conditioning seen by
an individual specialist, and therefore does not separately identify
the effects of routing, prompt length, and specialist identity.

\subsection{RL Post-Training Improves Integrative Judgment}

\begin{figure}[t]
    \centering
    \includegraphics[width=0.98\linewidth]{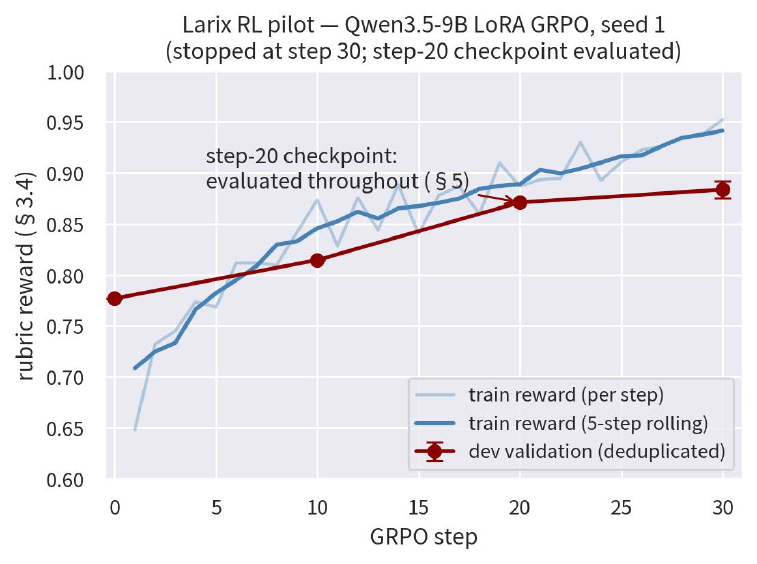}
    \caption{
    Training reward (blue, 5-step rolling) and in-training development
    validation (red, deduplicated over unique task--firm rows) over the
    30-update GRPO run of Qwen3.5-9B, stopped early for budget.
    The preserved step-20 checkpoint is evaluated throughout
    (Sec.~\ref{sec:experimental_setup}).
    }
    \label{fig:training_reward}
\end{figure}

GRPO post-training improves the development-split score of Qwen3.5-9B
from 72.2\% for Qwen-ZS to 84.2\% for Qwen-RL, an improvement of
12.0 percentage points (Fig.~\ref{fig:training_reward}); scores are
averaged over three dispatches at temperature 0.7 on the 25 unique
development rows. T1 is unchanged at its zero-shot ceiling (99.5\% in
both conditions); every other task improves. The judgment aggregate
(T2 and T5) increases by 14.2 points, from 58.6\% to 72.8\%.

The task-level development gains are 0.0 points on T1, +13.3 points
on T2 (80.0\% to 93.3\%), +18.9 points on T3 (79.8\% to 98.7\%),
+15.0 points on T5 (37.2\% to 52.2\%), and +12.4 points on T6
(64.6\% to 77.0\%). The judgment-task gains carry the paper's claim:
T2 and T5 are the tasks on which prompt-level specialist decomposition
provides no consistent advantage, and post-training improves both.
T3's gain partly recovers a zero-shot baseline depressed by
temperature-0.7 sampling: under greedy decoding the zero-shot model
already scores 98.7\% on T3, and the post-trained model matches it.
The same direction holds under greedy decoding, the protocol most
favorable to the zero-shot baseline: the overall gain is 3.4 points
(79.2\% to 82.6\%), the judgment-aggregate gain is 7.5 points
(64.2\% to 71.7\%), and T1 and T3 are unchanged. We report the
sampled protocol as primary because it matches the held-out evaluation
protocol (Sec.~\ref{sec:ood_transfer}).

The within-model comparison isolates the effect of post-training: the
architecture, specialist prompt, evidence, output schema, and scorer
are unchanged between Qwen-ZS and Qwen-RL. Two scope limits apply.
First, this 30-step training run evaluates one preserved checkpoint and
establishes feasibility but does not characterize convergence. Second,
the development split is disjoint from training but drawn from the
same firm and wrapper distribution; the out-of-distribution evidence
is Sec.~\ref{sec:ood_transfer}.

\subsection{Out-of-Distribution Transfer and Reward Diagnostics}
\label{sec:ood_transfer}

Held-out evaluation uses three dispatches at temperature 0.7 against
the identical zero-shot checkpoint, scored with the same rubric
(Table~\ref{tab:heldout}). The development and held-out evaluations
both use the preserved step-20 checkpoint under the identical
standalone protocol. Mean score increases on all three
anti-memorization splits: from 67.4\% to 82.6\% (+15.2 points) on the
35-row held-out-firm split, from 64.8\% to 69.1\% (+4.3) on the
25-row held-out-wrapper split, and from 93.1\% to 95.6\% (+2.5) on
the 45-row held-out-period split.

The task-level pattern is consistent with the development split: the
largest out-of-distribution gains occur on the judgment tasks. T5
improves by 40.4 points on unseen firms (38.1\% to 78.5\%) and is
flat on unseen wrappers (51.7\% in both conditions); T2 improves by
9.5 and 10.0 points respectively. The numerical tasks also transfer:
T1 gains 9.3 points on both unseen firms and unseen wrappers and 1.9
on later periods, T3 is stable (+1.0, $-0.3$, +4.2), and T6 gains
16.2 points on unseen firms but remains near floor on unseen wrappers
(24.4\% to 26.9\%), the one regime where neither checkpoint has
reliable signal. With 21 and 15 scored generations per task cell on
the firm and wrapper splits, movements under roughly 7 points are
directional; the headline gains---T5 on unseen firms, T1 and T2 on
both splits---exceed that threshold.

A final transfer check evaluates the zero-shot and step-20 checkpoints
on the 95-cell primary benchmark (19 firms $\times$ 5 tasks,
specialist prompts, the benchmark's strict cell scorer; $k=3$ at
temperature 0.7). Qwen-RL improves the judgment aggregate from 34.2\%
to 51.8\% (+17.6 points; T2 +8.8, T5 +26.3) while the numerical
aggregate is unchanged (30.4\%), confirming that the development-split
judgment gain is an artifact of neither the rubric's partial credit nor
the training prompt distribution. T1 and T3 sit at the floor in both
conditions (0.0--1.8\%): no-document extraction demands parametric
recall of FY2024 filings beyond an 8B model (the frontier model scores
84--100\% on identical prompts), so these cells measure model scale
rather than post-training. The document-bearing tasks (T2/T5/T6)
render under the v2 document budget and are not hash-identical to the
frozen frontier dispatches.

All held-out examples use real primary-source ground truth; the five
counterfactual breach-positive rows appear only during training.
Reward-term telemetry over all 15,360 training rollouts shows no
degenerate optimization of the $\beta$-weighted boolean term: the
breach-positive bonus fires on only 5.5\% of rollouts, ruling out an
``always predict breach'' collapse, and the dense continuous terms
saturate early (0.93--0.98), leaving the late-training gradient in the
threshold, unit, and calibrated-probability fields rather than the
breach-classification component alone.

\begin{table*}[t]
\centering
\small
\caption{
Held-out per-task scores, zero-shot Qwen3.5-9B (ZS) vs.\ the step-20
GRPO checkpoint (RL), in percent; $k=3$ dispatches at temperature
0.7 with the identical scorer, averaged over unique rows. Scores
increase on all three splits; the largest gains are the judgment tasks,
led by T5 on unseen firms (+40.4); T6 remains near floor on unseen
wrappers in both conditions.
}
\label{tab:heldout}
\setlength{\tabcolsep}{4pt}
\begin{tabular}{lccc ccc ccc}
\toprule
&
\multicolumn{3}{c}{Held-out firms}
&
\multicolumn{3}{c}{Held-out wrappers}
&
\multicolumn{3}{c}{Later periods}
\\
\cmidrule(lr){2-4}
\cmidrule(lr){5-7}
\cmidrule(lr){8-10}
Task
& ZS & RL & $\Delta$
& ZS & RL & $\Delta$
& ZS & RL & $\Delta$ \\
\midrule
T1 & 88.1 & 97.4 & +9.3
   & 75.7 & 85.0 & +9.3
   & 91.8 & 93.7 & +1.9 \\

T2 & 78.6 & 88.1 & +9.5
   & 73.3 & 83.3 & +10.0
   & -- & -- & -- \\

T3 & 80.5 & 81.5 & +1.0
   & 98.7 & 98.4 & $-0.3$
   & 95.8 & 100.0 & +4.2 \\

T5 & 38.1 & 78.5 & +40.4
   & 51.7 & 51.7 & 0.0
   & -- & -- & -- \\

T6 & 51.4 & 67.6 & +16.2
   & 24.4 & 26.9 & +2.5
   & -- & -- & -- \\
\midrule
All
& 67.4 & 82.6 & +15.2
& 64.8 & 69.1 & +4.3
& 93.1 & 95.6 & +2.5 \\
\bottomrule
\end{tabular}
\end{table*}

\section{Discussion and Limitations}

Specialist decomposition has a task-dependent effect. Scoped prompts
help localized numerical operations by removing unrelated instructions
and making intermediate fields explicit, but not judgments requiring
joint reasoning over accounting, regulatory, covenant, and
company-specific evidence; narrow prompt scope may omit cross-lens
evidence required for integrative decisions, although the present
experiments do not isolate this mechanism.

The full-framework control shows that additional domain instructions
alone do not reproduce the specialist gain, and GRPO post-training
instead changes the model's decision policy and improves the judgment
tasks, with gains that transfer to unseen firms and unseen regulatory
wrappers. Prompt decomposition and parameter adaptation thus address
distinct bottlenecks; the post-training result reflects task-specific
specialization, not general small-model superiority.

The evaluation is scoped to the specialist layer: 19 firms across
seven wrappers, three lens-aligned specialists, and five tasks, with
synthesis and position sizing outside the present design; extending
the benchmark across financial domains, wrappers, model families, and
agent topologies is natural future work. Per-wrapper estimates rest on
few firms, and the out-of-distribution transfer evidence comes from one
preserved checkpoint. Ground truth combines 45 expert-blinded tuples
with 50 author-curated continental tuples carrying per-cell provenance.
The RL corpus is compact (195 real dispatches plus five training-only
counterfactual breach-positive examples), and the post-training
comparison uses one frontier model family and one 9B open model.

\section{Conclusion}

We evaluated lens-aligned specialist decomposition and structured RL
post-training for regime-aware analysis of European listed real estate.
On the 19-firm, seven-wrapper primary benchmark, specialist
decomposition improves the numerical aggregate by 15.8 percentage
points but changes the judgment aggregate by only 5.3 points on average
across dispatches, with the judgment effect ranging from 0.0 to +10.5
across the four dispatches. A monolithic control given the complete
16-lens framework does not reproduce the numerical gain, indicating
that scoped lens assignment rather than framework disclosure alone
drives the improvement. GRPO post-training raises the development
score of Qwen3.5-9B by 12.0 points, with the judgment aggregate up
14.2 points and gains on all four sub-ceiling tasks; the gains transfer
to unseen firms (+15.2 points overall; +40.4 on covenant stress),
unseen regulatory wrappers (+4.3), and later periods (+2.5). A
frontier-model comparison on the expanded held-out splits remains
future work. Together, the results support a task-dependent design
principle: use decomposition to simplify localized, verifiable
financial operations, and use targeted parameter adaptation when
reliable performance depends on integrating multiple sources of
financial evidence. These conclusions remain bounded by the benchmark
size, wrapper coverage, model families, and the unevaluated downstream
synthesis layer.

\clearpage
\bibliographystyle{plainnat}
\bibliography{lorax_arxiv}

@article{ref1,
  author = {Barkham, R. and Ward, C.},
  title = {Investor Sentiment and Noise Traders: Discount to Net Asset Value in Listed Property Companies in the {U.K.}},
  journal = {Journal of Real Estate Research},
  volume = {18},
  number = {2},
  pages = {291--312},
  year = {1999}
}

@inproceedings{ref2,
  author = {Chen, Z. and Chen, W. and others},
  title = {{FinQA}: A Dataset of Numerical Reasoning over Financial Data},
  booktitle = {Proceedings of the 2021 Conference on Empirical Methods in Natural Language Processing (EMNLP)},
  year = {2021}
}

@article{ref3,
  author = {{DeepSeek-AI}},
  title = {{DeepSeek-R1}: Incentivizing Reasoning Capability in {LLMs} via Reinforcement Learning},
  journal = {arXiv preprint},
  year = {2025}
}

@misc{ref4,
  author = {{European Public Real Estate Association}},
  title = {{EPRA} Best Practices Recommendations ({BPR}) Guidelines},
  year = {2024},
  howpublished = {\url{https://www.epra.com/finance/epra-reporting-bpr-guidelines}}
}

@article{ref5,
  author = {Fatouros, G. and Metaxas, K. and Soldatos, J.},
  title = {{MarketSenseAI} 2.0: Enhancing Stock Analysis through {LLM} Agents},
  journal = {arXiv preprint},
  year = {2025}
}

@article{ref6,
  author = {Ghosh, Chinmoy and Petrova, Milena},
  title = {The Effect of Legal Environment and Regulatory Structure on Performance: Cross-Country Evidence from {REITs}},
  journal = {Journal of Real Estate Finance and Economics},
  year = {2021},
  note = {doi:10.1007/s11146-019-09742-8}
}

@inproceedings{ref7,
  author = {Guo, Taicheng and Chen, Xiuying and Wang, Yaqi and others},
  title = {Large Language Model Based Multi-Agents: A Survey of Progress and Challenges},
  booktitle = {Proceedings of the 33rd International Joint Conference on Artificial Intelligence (IJCAI)},
  year = {2024},
  note = {arXiv:2402.01680}
}

@article{ref8,
  author = {Hong, S. and Zhuge, M. and Chen, J. and others},
  title = {{MetaGPT}: Meta Programming for a Multi-Agent Collaborative Framework},
  journal = {arXiv preprint},
  year = {2023}
}

@article{ref9,
  author = {Islam, P. and Kannappan, A. and others},
  title = {{FinanceBench}: A New Benchmark for Financial Question Answering},
  journal = {arXiv preprint arXiv:2311.11944},
  year = {2023}
}

@article{ref10,
  author = {Ke, Q.},
  title = {Valuation of European-Listed Real Estate Company: An Accounting Approach as Alternative},
  journal = {Journal of European Real Estate Research},
  year = {2026},
  note = {doi:10.1108/JERER-10-2025-0085}
}

@article{ref11,
  author = {Liu, W. and others},
  title = {{Fin-R1}: A Large Language Model for Financial Reasoning through Reinforcement Learning},
  journal = {arXiv preprint},
  year = {2025}
}

@article{ref12,
  author = {Ouyang, L. and Wu, J. and Jiang, X. and others},
  title = {Training Language Models to Follow Instructions with Human Feedback},
  journal = {Advances in Neural Information Processing Systems (NeurIPS)},
  year = {2022}
}

@misc{ref13,
  author = {{Qwen Team}},
  title = {{Qwen3.5-9B} (Hugging Face Model Card)},
  year = {2026},
  howpublished = {\url{https://huggingface.co/Qwen/Qwen3.5-9B}}
}

@article{ref14,
  author = {Shao, Z. and Wang, P. and Zhu, Q. and others},
  title = {{DeepSeekMath}: Pushing the Limits of Mathematical Reasoning in Open Language Models},
  journal = {arXiv preprint},
  year = {2024}
}

@article{ref15,
  author = {Sheng, G. and Zhang, S. and others},
  title = {{HybridFlow}: A Flexible and Efficient {RL} Framework for {LLMs} ({veRL})},
  journal = {arXiv preprint},
  year = {2024}
}

@article{ref16,
  author = {Vincent, L.},
  title = {The Information Content of Funds from Operations ({FFO}) for Real Estate Investment Trusts ({REITs})},
  journal = {Journal of Accounting and Economics},
  volume = {26},
  number = {1--3},
  pages = {69--104},
  year = {1999}
}

@inproceedings{ref17,
  author = {Wei, J. and Wang, X. and Schuurmans, D. and others},
  title = {Chain-of-Thought Prompting Elicits Reasoning in Large Language Models},
  booktitle = {Advances in Neural Information Processing Systems 35 (NeurIPS)},
  year = {2022}
}

@article{ref18,
  author = {Wu, S. and Irsoy, O. and Lu, S. and others},
  title = {{BloombergGPT}: A Large Language Model for Finance},
  journal = {arXiv preprint},
  year = {2023}
}

@article{ref19,
  author = {Xiao, Yijia and Sun, Edward and Luo, Di and Wang, Wei},
  title = {{TradingAgents}: Multi-Agents {LLM} Financial Trading Framework},
  journal = {arXiv preprint arXiv:2412.20138},
  year = {2024}
}

@inproceedings{ref20,
  author = {Xie, Q. and Han, W. and Lai, Z. and others},
  title = {{FinBen}: A Holistic Financial Benchmark for Large Language Models},
  booktitle = {Advances in Neural Information Processing Systems (NeurIPS), Datasets and Benchmarks Track},
  year = {2024}
}

@article{ref21,
  author = {Yang, H. and Zhang, B. and Wang, N. and others},
  title = {{FinRobot}: An Open-Source {AI} Agent Platform for Financial Applications Using Large Language Models},
  journal = {arXiv preprint},
  year = {2024}
}

@inproceedings{ref22,
  author = {Yu, H. and others},
  title = {{FinMem}: A Performance-Enhanced {LLM} Trading Agent with Layered Memory and Character Design},
  booktitle = {Proceedings of the AAAI Conference on Artificial Intelligence},
  year = {2023}
}

@inproceedings{ref23,
  author = {Yu, Yangyang and Yao, Zhiyuan and Li, Haohang and others},
  title = {{FinCon}: A Synthesized {LLM} Multi-Agent System with Conceptual Verbal Reinforcement for Enhanced Financial Decision Making},
  booktitle = {Advances in Neural Information Processing Systems 37 (NeurIPS)},
  pages = {137010--137045},
  year = {2024},
  note = {arXiv:2407.06567}
}

@article{ref24,
  author = {Zhao, Tianjiao and Lyu, Jingrao and Jones, Stokes and others},
  title = {{AlphaAgents}: Large Language Model Based Multi-Agents for Equity Portfolio Constructions},
  journal = {arXiv preprint arXiv:2508.11152},
  year = {2025}
}

@article{ref25,
  author = {Zheng, Chujie and Liu, Shixuan and Li, Mingze and others},
  title = {Group Sequence Policy Optimization},
  journal = {arXiv preprint arXiv:2507.18071},
  year = {2025}
}

@article{ref26,
  author = {Zhu, J. and others},
  title = {{DianJin-R1}: Evaluating and Enhancing Financial Reasoning in Large Language Models},
  journal = {arXiv preprint},
  year = {2025}
}

@article{taghavi2026spec,
  title = {Spec Kit Agents: Context-Grounded Agentic Workflows},
  author = {Taghavi, Pardis and Bhavani, Santosh},
  journal = {arXiv preprint arXiv:2604.05278},
  year = {2026}
}

\clearpage
\appendix

\section{Supplementary Material}
\subsection{Harness-Hardening Findings}
\label{app:harness}

Two methodological findings emerged from hardening the evaluation
harness and generalize beyond this benchmark.

\paragraph{Field-request alignment.}
The evaluation prompt must explicitly request every structured field
the rubric scores. An earlier template variant asked only for surface
fields while the scorer graded structured sub-fields, producing
0\%-by-construction cells on the judgment tasks that reflected
prompt--scorer misalignment rather than model capability. Aligning the
schema, prompt, and scorer (Sec.~3.3) removes this artifact.

\paragraph{Rubric sub-classification.}
Rubrics for regime-dependent reasoning require sub-classification
rather than coarse categoricals. A four-way covenant categorical
saturated and could not differentiate conditions; the sub-classified
rubric (covenant type, unit, and threshold value under AND semantics)
restored discrimination. The same applies to free-text judgment fields,
where token-overlap scoring is fragile.

\paragraph{Numeric-tolerance sensitivity.}
The $\pm5\%$ (T1) and $\pm15\%$ (T3) numeric bars are the one
load-bearing scoring choice on the numerical tasks. Under the initial
dispatch, no T1 cell sat within $\pm1$ point of its bar, and the
borderline T3 cells left the T3 condition gap between +10.5 and +26.3
points under any tolerance in $[14\%,16\%]$, never changing sign.

\end{document}